\documentclass[11pt]{article}

\usepackage[preprint]{acl}

\usepackage{times}
\usepackage{latexsym}

\usepackage[T1]{fontenc}

\usepackage[utf8]{inputenc}

\usepackage{microtype}

\usepackage{inconsolata}

\usepackage{graphicx}

\usepackage{amsmath} 
\usepackage{amsfonts}
\usepackage{multirow} 
\usepackage{graphicx}
\usepackage{subcaption}

\title{Knowledge Distillation for Temporal Knowledge Graph Reasoning with Large Language Models}

\author{Wang Xing\textsuperscript{1},
Wei Song\textsuperscript{2}, 
Siyu Lin\textsuperscript{3},
Chen Wu\textsuperscript{2}, 
Zhesi Li\textsuperscript{4}, 
Man Wang\textsuperscript{2} \\
$^1$School of Computer Science and Technology, Xidian University\\ 
$^2$School of Computing and Artificial Intelligence, Southwest Jiaotong University\\
$^3$School of Information Science and Engineering, Chongqing Jiaotong University\\
$^4$School of Information Engineering, Chang'an University\\
}

\begin{document}
\maketitle
\begin{abstract}
Reasoning over temporal knowledge graphs (TKGs) is fundamental to improving the efficiency and reliability of intelligent decision-making systems and has become a key technological foundation for future artificial intelligence applications. Despite recent progress, existing TKG reasoning models typically rely on large parameter sizes and intensive computation, leading to high hardware costs and energy consumption. These constraints hinder their deployment on resource-constrained, low-power, and distributed platforms that require real-time inference. Moreover, most existing model compression and distillation techniques are designed for static knowledge graphs and fail to adequately capture the temporal dependencies inherent in TKGs, often resulting in degraded reasoning performance.
To address these challenges, we propose a distillation framework specifically tailored for temporal knowledge graph reasoning. Our approach leverages large language models as teacher models to guide the distillation process, enabling effective transfer of both structural and temporal reasoning capabilities to lightweight student models. By integrating large-scale public knowledge with task-specific temporal information, the proposed framework enhances the student model’s ability to model temporal dynamics while maintaining a compact and efficient architecture. Extensive experiments on multiple publicly available benchmark datasets demonstrate that our method consistently outperforms strong baselines, achieving a favorable trade-off between reasoning accuracy, computational efficiency, and practical deployability.

\end{abstract}

\section{Introduction}

A temporal knowledge graph (TKG) is a knowledge storage structure with explicit temporal attributes and has been widely applied in real world scenarios \citep{leblay2018deriving, kazemi2020representation}. With the rapid development of natural language processing, TKGs have become increasingly mature and have been extensively used in system operation, semantic matching, and operational scenario planning \citep{garcia2020temporal}. Representative TKGs include the YAGO knowledge base developed by the Max Planck Institute \citep{mahdisoltani2015yago3}, as well as the multilingual knowledge base WIKI proposed by the Wikimedia Foundation \citep{vrandevcic2014wikidata}.

Static knowledge graph reasoning models operate on event triples that do not contain temporal attributes, enabling models to understand and reason over knowledge graphs for tasks such as link prediction and path prediction \citep{bordes2013translating}. TransE is one of the pioneering methods in this field, mapping relationships as translations between entity embeddings \citep{bordes2013translating}. RotatE extends this idea by modeling relations as rotations in complex vector space, thereby enhancing representation capacity \citep{sun2019rotate}. ComplEx further encodes entities and relations in complex space \citep{trouillon2016complex}, while SimplE adopts embeddings in vector space to enrich entity and relation representations \citep{kazemi2018simple}. These methods have promoted the development of static knowledge graph reasoning and have also laid a solid foundation for temporal knowledge graph reasoning.

Temporal knowledge graph reasoning models knowledge in the form of translation vectors, representing the relationship between two entity embeddings as a translation vector \citep{leblay2018deriving}. Some approaches adopt deep neural network based methods, such as applying feedback mechanisms or convolutional layers in the embedding space \citep{jin2020recurrent}. However, these reasoning methods inherited from static knowledge graph research do not explicitly consider temporal attributes, which limits their ability to model the dynamic evolution of events.
Beyond embedding-based extensions of static knowledge graph models, 
recent work has explored incorporating explicit temporal logical structures 
to enhance reasoning over temporal knowledge graphs. 
Differentiable logical rule learning frameworks \citep{xiongtilp, xiong2024teilp} model temporal dependencies 
through logical rules and enable interpretable temporal reasoning and time prediction.

Temporal knowledge graph reasoning methods can be categorized into two different settings, namely interpolation and extrapolation \citep{xu2020temporal}. Given a TKG, interpolation based models aim to solve reasoning problems within a time interval $[t_0, t_N]$, which is also referred to as TKG completion \citep{leblay2018deriving}. In contrast, extrapolation based models focus on predicting events that may occur at future timestamps $t_{N+1}$ and beyond \citep{trivedi2017know}. Representative interpolation models include HyTE and related approaches \citep{dasgupta2018hyte}.

Chronos aims to infer missing entity relationships but cannot effectively capture temporal variation and struggles to generalize to unseen time periods \citep{sadeghian2021chronos}. Know-Evolve represents temporal evolution as a continuous process \citep{trivedi2017know}, while ExTTE provides interpretable reasoning and prediction, although its performance is affected by data sparsity \citep{xu2020temporal}. CyGNet introduces a cyclic evolution network to model periodic patterns in TKGs \citep{zhu2021cygnet}. Other methods employ graph neural networks or recurrent neural networks to capture temporal dynamics, such as RE-NET \citep{jin2020recurrent}.

Despite extensive research on TKG reasoning, many methods are designed for large scale knowledge graphs and rely on training models with massive parameters \citep{garcia2020temporal}. This results in high computational and storage requirements, which limits deployment on resource constrained devices. To address this issue, lightweight modeling and compression techniques have been explored \citep{hinton2015distilling}.

Model acceleration and compression have become active research topics. Knowledge distillation has been widely applied in areas such as computer vision and recommendation systems \citep{hinton2015distilling}. In the knowledge graph domain, distillation has been used to transfer knowledge from large models to compact student models. However, most existing distillation methods focus on static knowledge graphs and do not explicitly incorporate temporal information \citep{zhang2020relational}.

Large language models (LLMs), such as GPT-3 \citep{brown2020language}, PaLM \citep{chowdhery2022palm}, and LLaMA \citep{touvron2023llama}, demonstrate strong reasoning and generalization capabilities due to large scale pretraining. 
Recent advances, including instruction tuning and reinforcement learning from human feedback, further enhance their reasoning ability \citep{ouyang2022training}. 
Recent studies further show that LLMs can acquire non-trivial reasoning abilities and generalize to complex real-world reasoning scenarios \citep{xiong2024large, yang2024can}.
Beyond their raw reasoning capability, LLMs have also been shown to support structured reasoning and performing deliberate planning with internal world models \citep{yang2024harnessing, xiong2025deliberate}. 
These properties make LLMs promising teacher models for distillation.

Motivated by these observations, we propose a distillation method tailored for temporal knowledge graph reasoning. Our approach leverages large language models as teachers to integrate both public knowledge and task specific temporal knowledge into the distillation process. By guiding lightweight student models with rich temporal reasoning signals, the proposed method enables efficient deployment on low power distributed devices. Experimental results demonstrate that our approach consistently outperforms traditional TKG reasoning models while significantly reducing model size and computational cost.

\begin{figure*}[t]
\centering
\begin{subfigure}[t]{0.4\textwidth}
  \centering
  \includegraphics[width=\linewidth]{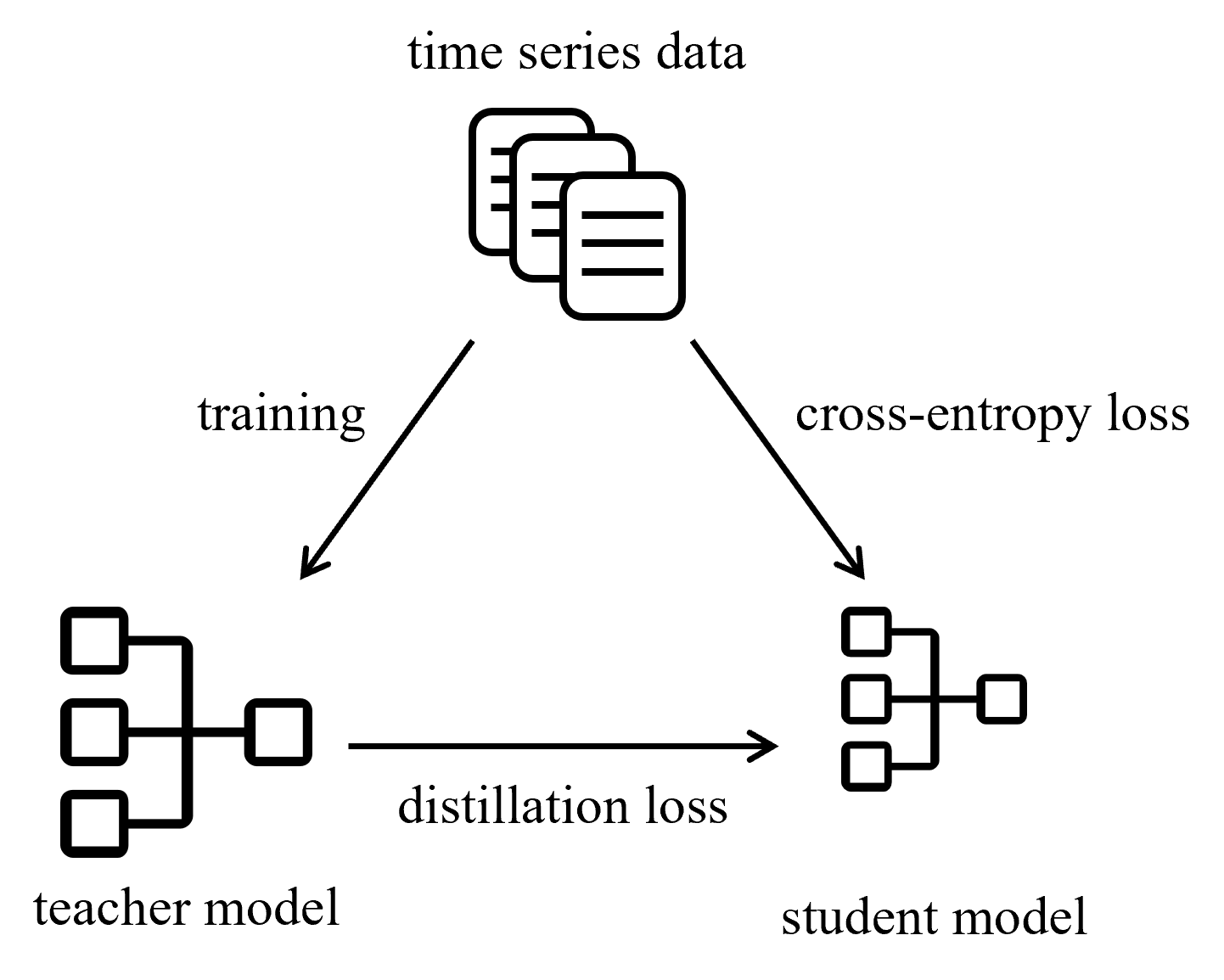}
  \caption{Traditional knowledge distillation method}
\end{subfigure}
\hspace{2pt}
\begin{subfigure}[t]{0.4\textwidth}
  \centering
  \includegraphics[width=\linewidth]{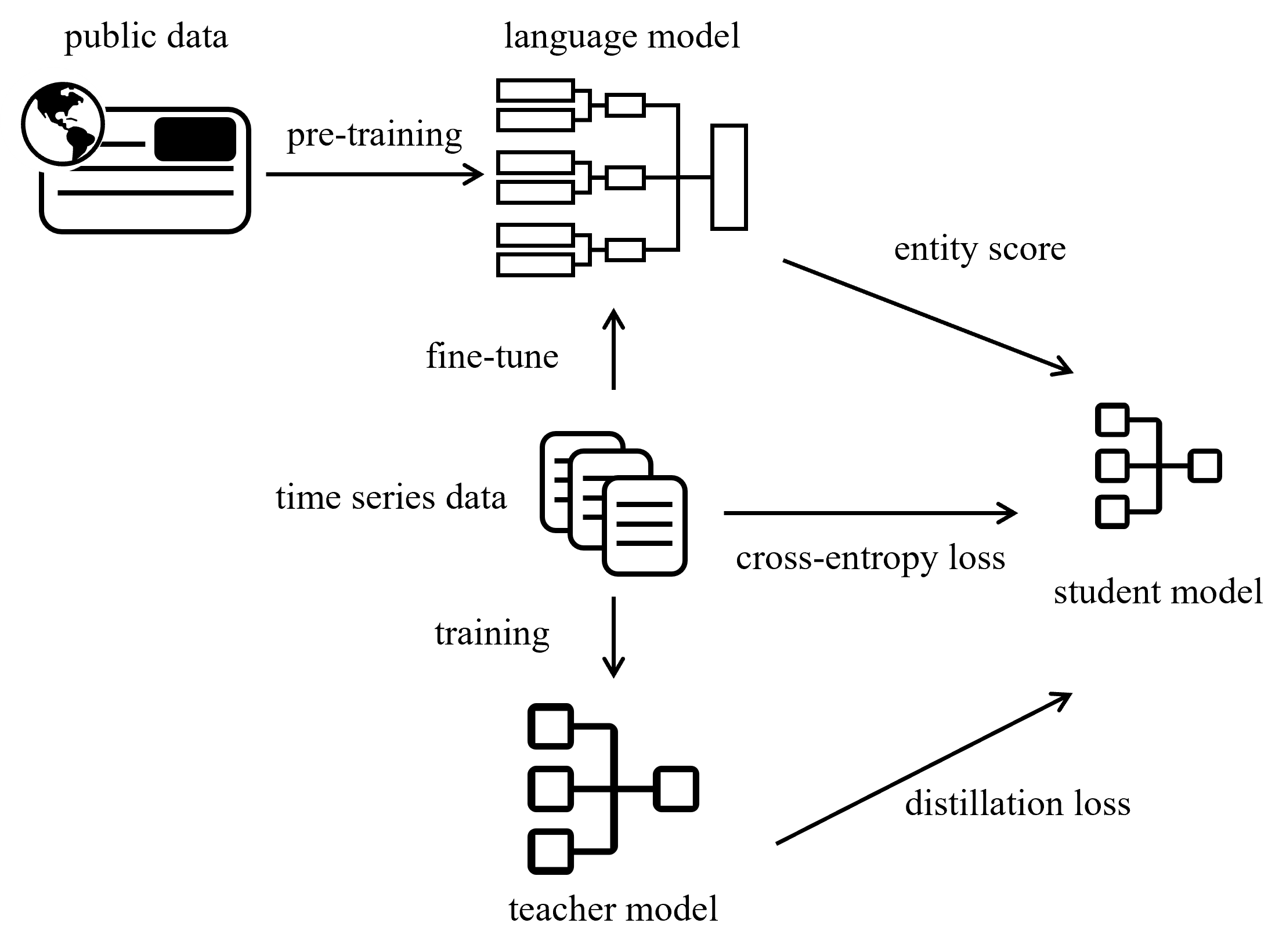}
  \caption{Large language model based distillation method}
\end{subfigure}
\caption{Comparison between different distillation methods.} 
\label{fig:distillation_comparison}
\end{figure*}

\section{Distillation Method for Temporal Knowledge Graph Reasoning Models}

\subsection{Problem Definition}

A temporal knowledge graph is defined as a set of temporal fact quadruples $(s, p, o, t)$, where $s$ denotes the subject entity, $p$ denotes the relation, $o$ denotes the object entity, and $t$ denotes the timestamp. Figure 1 illustrates an example of a national relationship evolution process, where a sequence of events occurs at different timestamps. Some entity events occur multiple times at different time points, while others occur only once, forming a dynamic temporal structure.

Given a TKG $\mathcal{G}$, the objective of temporal knowledge graph reasoning is to predict missing facts at a given timestamp or to forecast future events based on historical observations. This task is commonly referred to as temporal knowledge graph reasoning and can be formulated as a link prediction problem under temporal constraints.

Future temporal prediction aims to answer queries such as what the relationship between Ukraine and the United States will be at a future time $t_4$.

Unlike conventional knowledge graphs, temporal knowledge graphs explicitly incorporate temporal information. In a TKG, each fact is represented as a quadruple $(s, p, o, t)$, where $s \in E$ denotes the subject entity, $o \in E$ denotes the object entity, $p \in R$ denotes the relation, and $t \in T$ denotes the timestamp. Here, $E$ and $R$ represent the sets of entities and relations, respectively, and $T$ denotes the set of timestamps. Let $\mathcal{G}_t$ denote the TKG at time $t$, and let $g = (s, p, o, t)$ represent a quadruple fact in $\mathcal{G}_t$. A TKG is a sequence of fact sets ordered by time, denoted as $\mathcal{G} = \{\mathcal{G}_1, \mathcal{G}_2, \ldots, \mathcal{G}_T\}$, where each $\mathcal{G}_t$ contains all facts occurring at time $t$.

Temporal knowledge graph reasoning focuses on predicting missing facts within the graph. Specifically, given a set of observed facts, the model aims to infer missing object entities $(s, p, ?, t)$ or missing subject entities $(?, p, o, t)$. In the experimental setting of this work, the reasoning model is applied to temporal fact prediction tasks, where either the subject or object entity is missing.

\begin{figure*}[t]
  \centering
  \includegraphics[width=0.9\linewidth]{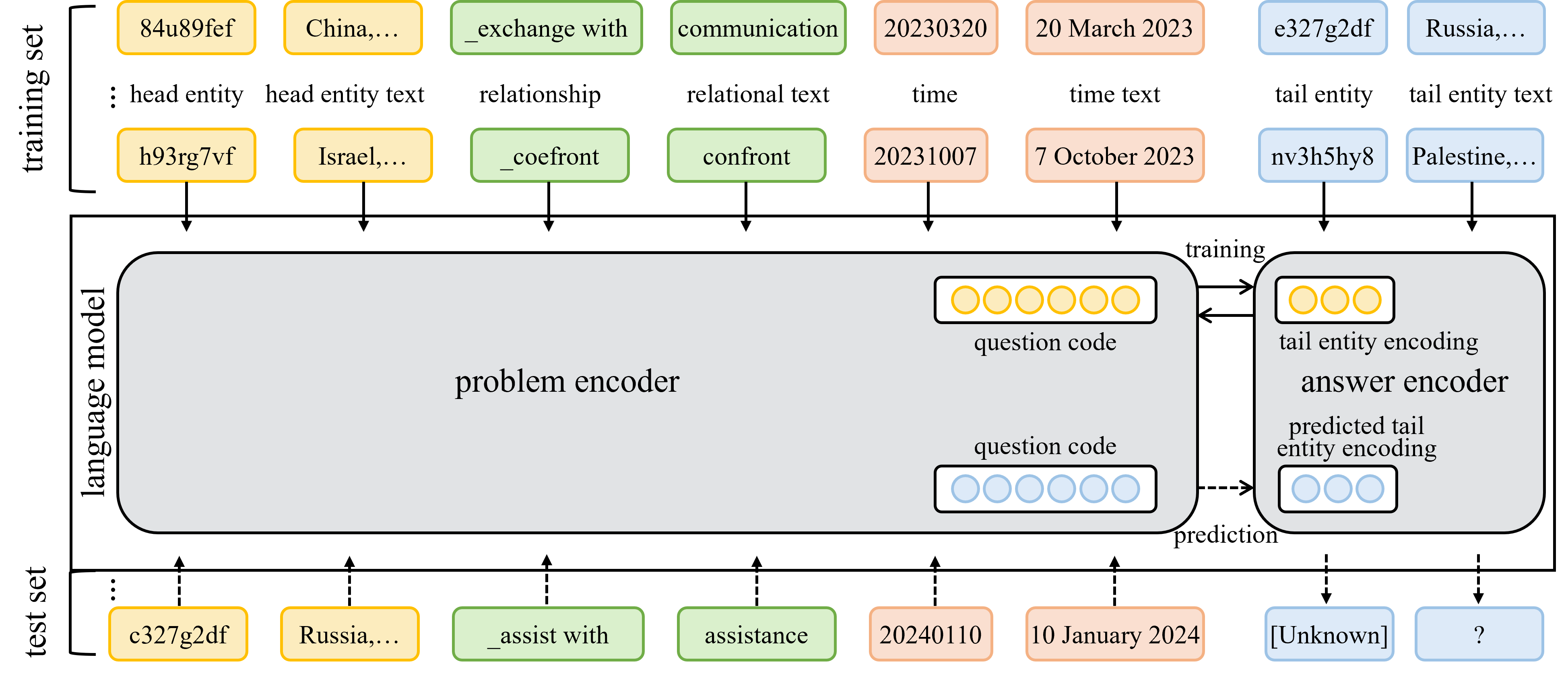}
  \caption{The training and prediction on quadruple knowledge with the large language model.}
  \label{fig:framework}
\end{figure*}

At time $t$, the TKG reasoning task constructs a set of missing fact quadruples $\mathcal{G}'_t$, where each quadruple is generated by randomly replacing the subject entity or object entity in the original fact set $\mathcal{G}_t$. These two types of incomplete fact sets together form the complete temporal knowledge graph $\mathcal{G}_c$, defined as
\begin{equation}
\mathcal{G}_c = \mathcal{G} \cup \mathcal{G}' .
\end{equation}

Formally, the set of incomplete facts is defined as
\begin{equation}
\begin{aligned}
\mathcal{G}' =\;&
\big\{(s', p, o, t) \mid s' \in E, s' \neq s\big\} \\
&\cup
\big\{(s, p, o', t) \mid o' \in E, o' \neq o\big\}.
\end{aligned}
\end{equation}

\subsection{Model Overview}

The proposed distillation framework for temporal knowledge graph reasoning based on large language models is inspired by traditional knowledge distillation methods. In the distillation process, a high capacity reasoning model is selected as the teacher model, while a lightweight model is trained as the student model. The teacher model is first pretrained on temporal data, after which the student model is optimized by minimizing distillation losses derived from the teacher.

In addition to the teacher model, a large language model is introduced to enhance the distillation process. The large language model is pretrained on large scale public datasets and further fine tuned on the temporal knowledge graph data, enabling it to serve as a secondary teacher model. During training, the student model jointly learns from both the traditional teacher model and the large language model.

During inference, the teacher model, large language model, and student model can be evaluated independently. To better align the prediction distributions between the student model and the teacher models, a two stage distillation strategy is adopted. The first stage focuses on aligning the student model with the traditional teacher model, while the second stage leverages the predictive distributions produced by the large language model to further refine the student model.

\subsection{Loss Functions}

During training, three types of loss functions are introduced. The final objective function is defined as a weighted sum of these losses. Let $L_1$ denote the distillation loss between the student model and the traditional teacher model, $L_2$ denote the distillation loss between the student model and the large language model, and $L_3$ denote the supervised loss based on ground truth labels. The total loss is defined as
\begin{equation}
L_{\text{total}} = L_1 + L_2 + \beta L_3 .
\end{equation}

The first loss function $L_1$ is designed to transfer temporal reasoning knowledge from the traditional teacher model to the student model. An encoder decoder architecture is adopted to align the hidden representations produced by the two models. Given a fact quadruple $(s, p, o, t)$, the distillation loss is defined as
\begin{align}
L_1 =\;& \alpha \, \mathcal{C}\big(g \log f_T(s, p, o, t)\big) \nonumber \\
& + (1 - \alpha)\, \mathcal{C}\big(g \log f_S(s, p, o, t)\big).
\end{align}
where $f_T(\cdot)$ and $f_S(\cdot)$ denote the scoring functions of the teacher model and the student model, respectively, $\alpha$ is a balancing coefficient, and $\mathcal{C}(\cdot)$ denotes the SoftMax function.

To reduce the discrepancy between the prediction distributions of the student model and the large language model, the second loss function $L_2$ adopts the Huber loss, which provides robustness to outliers while maintaining sensitivity to small errors. The loss is defined as
\begin{equation}
L_2 =
\begin{cases}
\frac{1}{2} (f_T - f_S)^2, & |f_T - f_S| \le \delta, \\
\delta |f_T - f_S| - \frac{1}{2} \delta^2, & |f_T - f_S| > \delta,
\end{cases}
\end{equation}
where $f_T$ and $f_S$ represent the prediction scores of the large language model and the student model, respectively, and $\delta$ is a threshold hyperparameter.

The third loss function $L_3$ is the supervised loss based on ground truth labels. The large language model is used to encode entity relation semantics, producing a dense vector representation. The prediction process is defined as
\begin{equation}
\mathbf{E}(p) = \sigma\big(w_p \, \text{LLM}(e) + b_p\big),
\end{equation}
\begin{equation}
P(p) = \sigma\big(w \cdot \mathbf{E}(p) + b\big),
\end{equation}
where $\text{LLM}(\cdot)$ denotes the output embedding of the large language model, $w_p$, $b_p$, $w$, and $b$ are learnable parameters, and $\sigma$ denotes the SoftMax function.

Finally, the supervised loss is computed using mean squared error
\begin{equation}
L_3 = \| f_S(p) - P(p) \|_2^2 .
\end{equation}

The output of the large language model is represented as a vector.

The scoring function of the large language model is incorporated into the distillation process through a soft distillation loss, which is computed using mean squared error
\begin{equation}
L^{\text{LLM}}_3 = \left\| f^{\text{LLM}}(E(s), P(p), E(o), t) - f^S \right\|_2^2 .
\end{equation}
Here, $f^{\text{LLM}}$ denotes the scoring function of the large language model.

The final loss function used to train the student model is defined as a weighted combination of the above three loss terms
\begin{equation}
L_{\text{total}} = L_1 + \alpha L_2 + \beta L^{\text{LLM}}_3 .
\end{equation}

\section{Experimental Setup}

This section presents the experimental setup, including the datasets, baseline models, evaluation metrics, and training configurations.

\subsection{Datasets}

We select two publicly available benchmark datasets in the temporal knowledge graph domain, namely YAGO11k and WIKIdata12k, which are widely used for evaluating temporal knowledge graph reasoning models \citep{leblay2018deriving, garcia2020temporal}. These datasets contain temporal events annotated with timestamps and support both interpolation and extrapolation settings.

The YAGO dataset is derived from the knowledge base developed by the Max Planck Institute and consists of factual data collected from Wikipedia, WordNet, and GeoNames \citep{mahdisoltani2015yago3}. It includes rich temporal and relational information, making it suitable for evaluating temporal reasoning performance. The dataset used in our experiments is YAGO11k.

The WIKI dataset is constructed from Wikidata and Wikimedia Commons, where Wikidata serves as the structured knowledge base for Wikipedia \citep{vrandevcic2014wikidata}. It contains a large number of multilingual facts and temporal annotations. The dataset used for temporal knowledge graph reasoning in our experiments is WIKIdata12k.

\subsection{Baseline Models}

BKD is a classical knowledge distillation method that minimizes the Kullback Leibler divergence between the output distributions of the teacher model and the student model \citep{hinton2015distilling}. It serves as the baseline distillation approach.

FitNet extends BKD by introducing intermediate layer supervision, enabling the student model to learn representations from multiple layers of the teacher network and thereby providing richer training signals \citep{romero2015fitnets}.

RKD is a structure based distillation method that transfers relational knowledge by matching the distances and angles between embedding vectors of the teacher and student models, explicitly capturing relational factors among samples.

\begin{table}[t]
\centering
\caption{Statistical data of two datasets}
\label{tab:dataset_statistics}
\resizebox{\linewidth}{!}{
\begin{tabular}{lccccc}
\hline
Dataset & Entities & Relations & Train & Valid & Test \\
\hline
YAGO & 10{,}623 & 10 & 161{,}540 & 19{,}523 & 20{,}026 \\
WIKI & 12{,}544 & 24 & 539{,}286 & 67{,}538 & 63{,}110 \\
\hline
\end{tabular}
}
\end{table}

\begin{table}[t]
\centering
\caption{Introduction of classic TKG models}
\label{tab:tkg_models}
\resizebox{\linewidth}{!}{
\begin{tabular}{lcc}
\hline
Model & Encoder & Decoder \\
\hline
TTransE & -- & $\lVert s + p - o + t \rVert_2$ \\
TADistMult & $p_{\text{seq}} = \mathrm{LSTM}(p:t)$ & $(s \circ o)\, p_{\text{seq}}^{\top}$ \\
\hline
\end{tabular}
}
\end{table}

\subsection{Model Selection}

The selected knowledge graph embedding models are TTransE and TADistMult. These models are representative temporal knowledge graph reasoning approaches and have demonstrated strong performance on temporal link prediction tasks \citep{leblay2018deriving, garcia2020temporal}. The encoder and decoder configurations of these models follow their original implementations.

In these models, $s, o \in \mathbb{R}^d$ denote the embeddings of subject and object entities, respectively, while $p \in \mathbb{R}^d$ represents the embedding of the relation type. The variable $t$ denotes the temporal embedding, and the operator $\circ$ denotes element wise multiplication.

\begin{table*}[t]
\centering
\caption{Distillation results of different model method combinations on two datasets}
\label{tab:distillation_results}
\setlength{\tabcolsep}{4pt}
\resizebox{\textwidth}{!}{
\begin{tabular}{llccccc|ccccc}
\hline
 &  & \multicolumn{5}{c|}{YAGO} & \multicolumn{5}{c}{WIKI} \\
\cline{3-12}
Model & Method
& MRR & MR & Hits@1 & Hits@3 & Hits@10
& MRR & MR & Hits@1 & Hits@3 & Hits@10 \\
\hline
\multirow{4}{*}{TTransE}
& BKD
& \underline{7.65} & 1410.12 & 3.50 & \underline{7.83} & 15.61
& \textbf{7.94} & 2383.67 & \underline{4.75} & \underline{8.22} & 14.04 \\
& FitNet
& 7.59 & 1201.69 & 3.06 & 7.18 & \underline{16.48}
& 7.86 & 2148.86 & 3.93 & 7.78 & \underline{14.67} \\
& RKD
& 7.01 & \textbf{1186.27} & \underline{3.56} & 6.95 & 13.47
& 7.89 & \underline{2052.37} & 4.72 & 7.49 & 12.85 \\
& Ours
& \textbf{7.69} & \underline{1193.15} & \textbf{3.61} & \textbf{7.89} & \textbf{16.57}
& \underline{7.92} & \textbf{1985.63} & \textbf{4.86} & \textbf{8.36} & \textbf{14.94} \\
\hline
\multirow{4}{*}{TADistMult}
& BKD
& \textbf{61.90} & \underline{973.89} & \underline{58.51} & \underline{64.13} & \underline{67.59}
& 45.89 & 3150.11 & \underline{42.46} & 48.87 & 51.18 \\
& FitNet
& 58.44 & 986.92 & 54.71 & 60.29 & 65.34
& 43.92 & 3158.20 & 39.77 & 47.38 & 50.18 \\
& RKD
& 58.15 & 1089.57 & 54.48 & 61.72 & 65.17
& 42.72 & 3287.49 & 36.32 & 43.92 & 47.28 \\
& Ours
& \underline{61.87} & \textbf{965.35} & \textbf{58.73} & \textbf{64.15} & \textbf{67.68}
& \textbf{46.03} & 3142.85 & 42.50 & \textbf{49.16} & \underline{51.14} \\
\hline
\end{tabular}
}
\end{table*}

\subsection{Evaluation Metrics}

We use two publicly available datasets for training and evaluation. The reported metrics include mean rank (MR), mean reciprocal rank (MRR), and top k ranking accuracy, namely Hits at 1, Hits at 3, and Hits at 10, which are standard evaluation metrics for knowledge graph reasoning and link prediction tasks \citep{bordes2013translating, trouillon2016complex}. Hits at k measures the proportion of test queries for which the correct entity is ranked within the top k positions in the prediction list.

For all metrics, a lower mean rank indicates better performance, while higher values of mean reciprocal rank and Hits at k reflect stronger reasoning capability \citep{sun2019rotate}.

\subsection{Training Settings}

All experiments are conducted on a machine equipped with an Intel Core i9 10900K CPU and an NVIDIA GeForce RTX 3090 Ti GPU. The models are implemented based on an extended version of the OpenKE framework \citep{han2018openke}, which is built upon PyTorch.

To demonstrate the effectiveness of the proposed method under significant capacity gaps between teacher and student models, the embedding dimension of the teacher model is set to 400, while the embedding dimension of the student model is set to 25. This configuration follows standard settings in prior work on knowledge distillation and lightweight knowledge graph embedding models \citep{hinton2015distilling}.

During training, the batch size is set to 1024, and the maximum number of training epochs is set to 10000 to ensure full convergence. In the distillation process, the temperature parameter is set to 7, following common practice in knowledge distillation \citep{hinton2015distilling}. The Adagrad optimizer is adopted to handle sparse data and adaptively adjust the learning rate \citep{duchi2011adaptive}.

\section{Results and Analysis}

\subsection{Experimental Results}

The effectiveness of the proposed distillation method is evaluated on the public YAGO and WIKI datasets. The experimental results under different model and method configurations are reported for both datasets. Except for the mean rank metric, all reported results are expressed as percentages. Under the same experimental settings, the best results are highlighted, while the second best results are underlined.

As shown by the experimental results, the student models trained with the proposed distillation method consistently outperform those trained with traditional distillation approaches. Compared with the baseline BKD method, the proposed approach achieves improvements on almost all evaluation metrics. For the TTransE model on the YAGO dataset, performance gains of 0.5 percent, 15.4 percent, 3.1 percent, 0.8 percent, and 6.1 percent are observed across the five evaluation metrics. Similarly, on the WIKI dataset, improvements of 16.7 percent, 2.3 percent, 1.7 percent, and 6.4 percent are achieved on mean rank, Hits at 1, Hits at 3, and Hits at 10, respectively.

Consistent performance improvements are also observed for the TADistMult model. These results demonstrate that the proposed distillation framework is effective and more capable of transferring knowledge to lightweight student models than conventional distillation methods.

By comparing the proposed approach with two classical knowledge distillation methods, namely FitNet and RKD, we observe that the proposed method achieves stable and superior performance across both public datasets. In particular, for the TADistMult model, the proposed method outperforms FitNet on all five evaluation metrics. The average improvements reach 2.77 percent on the YAGO dataset and 3.28 percent on the WIKI dataset, which further validates the effectiveness of the proposed approach.

It is worth noting that the proposed method achieves the best performance on Hits at 1 and Hits at 3. This indicates that the student models trained with the proposed framework are more effective at ranking correct entities at higher positions, which is critical for temporal knowledge graph reasoning. These results collectively demonstrate the advantage of the proposed method in addressing the challenges of distilling temporal knowledge graph reasoning models.

In a small number of experimental settings, the proposed method does not achieve the best performance on certain metrics. For example, when evaluating the TTransE model on the YAGO dataset, the distillation performance on mean rank is slightly inferior to the RKD baseline. Similarly, on the WIKI dataset, the performance of RKD exhibits noticeable fluctuations. This instability may be attributed to differences in the pretrained representations produced by large language models, which can make it difficult to effectively transfer knowledge under certain evaluation metrics. In addition, the relatively simple structure of student models may also contribute to this behavior. Experimental results suggest that increasing the number of training epochs can help alleviate this issue.

\subsection{Ablation Analysis}

We further conduct an ablation analysis based on the experimental results to evaluate the impact of large language model based knowledge distillation on student model training. By comparing the proposed method with the BKD baseline, we verify the effectiveness of incorporating large language models into the distillation framework.

On the YAGO dataset, when using TTransE as the backbone model, the proposed method improves performance on mean reciprocal rank, mean rank, Hits at 1, Hits at 3, and Hits at 10 by 0.5 percent, 15.4 percent, 3.1 percent, 0.8 percent, and 6.1 percent, respectively. For the TADistMult model, improvements of 7.0 percent, 4.0 percent, 1.0 percent, and 0.2 percent are observed on the corresponding metrics.

On the WIKI dataset, the student models trained with the proposed distillation method also achieve consistent improvements across most evaluation metrics compared with the BKD baseline. These results confirm that the knowledge injected from large language models effectively enhances the training of lightweight student models.

\section{Conclusion}

This paper proposes a distillation framework for temporal knowledge graph reasoning based on large language models, targeting efficient deployment on resource constrained devices. By leveraging the reasoning and knowledge integration capabilities of large language models, the proposed method effectively transfers temporal reasoning knowledge to lightweight student models. Experimental results across multiple datasets and backbone models demonstrate consistent improvements in reasoning accuracy while significantly reducing computational and storage requirements, outperforming existing distillation based approaches.

\section*{Limitations}

Despite its effectiveness, the proposed framework has several limitations. First, the quality of distillation depends on the availability and capability of the large language model, which may introduce additional computational overhead during training. Second, the current framework focuses on temporal link prediction and does not explicitly address more complex reasoning tasks such as multi hop temporal reasoning or logical rule induction. Finally, performance gains may vary across different backbone models and datasets, particularly when student models have extremely limited capacity. Addressing these limitations and extending the framework to broader reasoning scenarios remain important directions for future work.

\bibliography{custom}

\end{document}